\documentclass{article}

\usepackage[preprint]{manu}

\usepackage[utf8]{inputenc} 
\usepackage[T1]{fontenc}    
\usepackage{hyperref}       
\usepackage{url}            
\usepackage{booktabs}       
\usepackage{amsfonts}       
\usepackage{nicefrac}       
\usepackage{microtype}      
\usepackage{xcolor}         

\usepackage{natbib}
\setcitestyle{square,numbers,comma,sort&compress}

\usepackage{multirow}
\usepackage{array}
\usepackage{graphicx}
\usepackage{rotating} 
\usepackage{makecell} 
\usepackage{amsmath}
\usepackage{subcaption}
\definecolor{myblue}{HTML}{4682B4}  
\definecolor{mygreen}{HTML}{90EE90} 

\title{Context-Aware Probabilistic Modeling with LLM for Multimodal Time Series Forecasting}

\author{
    \textbf{Yueyang Yao}$^{1,2}$ \ \ \ \ 
    \textbf{Jiajun Li}$^{2}$ \ \ \ \ 
    \textbf{Xingyuan Dai}$^{1}$ \ \ \ \ 
    \textbf{Mengmeng Zhang}$^{1}$ \\
    \textbf{Xiaoyan Gong}$^{1}$ \ \ \ \
    \textbf{Fei-Yue Wang}$^{1}$ \ \ \ \ 
    \textbf{Yisheng Lv}$^{1}$\thanks{Corresponding author.} \\ 
    \\
    $^{1}$Institute of Automation, Chinese Academy of Sciences, China \\
    $^{2}$University of Chinese Academy of Sciences, China \\
}

\begin{document}

\maketitle

\begin{abstract}
  Time series forecasting is important for applications spanning energy markets, climate analysis, and traffic management. However, existing methods struggle to effectively integrate exogenous texts and align them with the probabilistic nature of large language models (LLMs). Current approaches either employ shallow text-time series fusion via basic prompts or rely on deterministic numerical decoding that conflict with LLMs' token-generation paradigm, which limits contextual awareness and distribution modeling. To address these limitations, we propose CAPTime, a context-aware probabilistic multimodal time series forecasting method that leverages text-informed abstraction and autoregressive LLM decoding. Our method first encodes temporal patterns using a pretrained time series encoder, then aligns them with textual contexts via learnable interactions to produce joint multimodal representations. By combining a mixture of distribution experts with frozen LLMs, we enable context-aware probabilistic forecasting while preserving LLMs’ inherent distribution modeling capabilities. Experiments on diverse time series forecasting tasks demonstrate the superior accuracy and generalization of CAPTime, particularly in multimodal scenarios. Additional analysis highlights its robustness in data-scarce scenarios through hybrid probabilistic decoding.  
\end{abstract}

\section{Introduction}
\label{sec:intro}
Time series forecasting plays a crucial role in diverse real-world applications, ranging from traffic flow prediction \cite{lvTrafficFlowPrediction2014} and weather forecasting \cite{wu2021autoformer} to energy price dynamics \cite{liu2024timemmd}. The integration of multimodal data, particularly textual information from reports and news, represents an important task with broad applications~\cite{jia2024gpt4mts,rodrigues2019combining}. For instance, beyond historical stock prices, analysts can derive accurate predictions by incorporating news sentiment and event impacts~\cite{dong2024fnspid}. Such exogenous text provides rich contextual information including causal explanations, event narratives, and market sentiment, thereby facilitating analysis of complex events and improving domain generalization.

The inherent complexity across domains necessitates a general forecasting model capable of handling diverse frequencies, multi-scale predictions and strong generalization abilities including zero-shot learning~\cite{bommasani2021foundation,das2024timesfm} for various applications. Currently, the development of general models suffers from limited pretraining data, and existing foundation models lack sufficient generalization. However, building on the similar sequential nature of time series and text, recent studies have successfully adapted large language models (LLMs) for time series forecasting~\cite{zhou2023gpt4ts,jintimellm,liu2024autotimes,hu2025context}, leveraging their pre-trained generalization capabilities from extensive textual corpora. Utilizing pre-trained LLMs~\cite{touvron2023llama,yang2024qwen2}, these approaches require minimal fine-tuning to achieve competitive performance.

Current LLM-based methods primarily address unimodal forecasting, achieving promising results by incorporating dataset statistics and descriptions through prompts. Existing methods face two challenges: (1) Few studies explore multimodal forecasting that leverages LLMs' native text understanding capabilities with exogenous textual data, which remains a promising research direction. Existing methods~\cite{caotempo,wang2024news,liu2024timemmd} either process exogenous texts as simple prompts or combine unimodal predictions through gating mechanisms, lacking deep modal fusion. (2) These methods utilize straightforward decoding approaches to transform discrete LLM outputs into continuous time series predictions. This paradigm differs fundamentally from LLMs' next-token distribution modeling objective, potentially limiting output diversity and representation capacity~\cite{rasul2023lagllama}.

To address these challenges, we propose \textbf{CAPTime}, a \textbf{C}ontext-\textbf{A}ware \textbf{P}robabilistic multimodal \textbf{Time} series forecasting method that extracts abstraction from exogenous text and performs forecasting via hybrid probabilistic decoding. Our approach enables both temporally aligned multimodal embedding and diverse distribution modeling through mixture of experts (MoE). Leveraging a pre-trained time series encoder, we derive universal temporal representations with high-level semantics. The proposed text abstraction module captures semantically relevant information for each temporal token through learnable interactions, generating text-informed multimodal token embeddings. To fully exploit LLM's token understanding and inter-token interaction capabilities, we maintain frozen parameters while employing probabilistic forecasting objectives to enhance LLMs' next-token distribution modeling. By integrating text abstraction with mixture of distribution experts, we achieve context-aware hybrid probabilistic decoding. This context-sensitive approach enables adaptive distribution modeling based on exogenous textual information, yielding more accurate and robust multimodal forecasting.

Our main contributions are summarized as follows:
\begin{itemize}
    \item We propose CAPTime, a context-aware framework for multimodal time series forecasting, which introduces a novel text abstraction mechanism to enhance multimodal token embeddings with richer semantics while enabling probabilistic forecasting.

    \item We introduce mixture of distribution experts to enhance the LLM's generalization capabilities in forecasting tasks, while obtaining universal temporal encoding through a pre-trained time series encoder.

    \item We conduct comprehensive evaluation of the proposed method across multiple forecasting tasks. Experimental results demonstrate the superior performance of our method, particularly for multimodal forecasting. Additional experiments reveal the method's strong few-shot and zero-shot generalization capabilities, along with enhanced forecasting performance through contextual text integration.
\end{itemize}

\section{Related Work}
\subsection{Time Series Analysis}
Time series analysis tasks involve forecasting, imputation, classification, and generation. These tasks are essential in domains like financial analysis and traffic management. Classical methods, such as ARIMA~\cite{box2015time} and SVR~\cite{sapankevych2009time}, model temporal dependencies directly. With the advent of deep learning, neural network-based forecasting methods have gained prominence \cite{wen2023transformers,zhou2021informer,wu2021autoformer,zeng2023dlinear}. Approaches based on RNNs or LSTMs capture sequential patterns through recurrent structures \cite{hochreiter1997long,qin2017dual,lai2018modeling}. Given the multivariate nature of time series, some studies employ CNNs to model both intra-variable and inter-variable dependencies \cite{bai2018empirical,wutimesnet,luo2024moderntcn}. More recently, transformer-based architectures have been adopted to model long-range dependencies and variable correlations \cite{zhou2022fedformer,liu2022non,challu2023nhits,nie2022time,liuitransformer}. However, these methods exhibit limitations in long-term complex forecasting, demonstrate weak zero-shot performance, and fail to effectively integrate exogenous information such as traffic news.

\subsection{Large Language Models for Time Series Forecasting}
Large language models have demonstrated strong natural language understanding across domains. Recent studies have explored LLM applications in time series forecasting, achieving promising results through their strong generalization\cite{liu2024timecma,hu2025context}. Initial approaches employed direct numerical-to-textual mapping of time series \cite{xue2023promptcast}. For example, LLMTIME \cite{gruver2023llmtime} converts time series values into text representations, demonstrating LLMs' potential as zero-shot time series forecasters. GPT4TS \cite{zhou2023gpt4ts} improves forecasting by aligning time series patches with embeddings and implementing partial fine-tuning. However, these methods rely on shallow modality alignment, limiting LLMs' capacity to effectively model temporal patterns and resulting in modest performance improvements.

Alternative approaches utilize textual prototypes or prompts to represent time series features \cite{liu2024unitime,suntest,liu2024autotimes,liu2024calf}. Time-LLM \cite{jintimellm} adapts LLMs by transforming time series into word embeddings. S\textsuperscript{2}IP-LLM \cite{pan2024s2ip} aligns textual and time series modalities in a latent space, employing semantic information as prompts. However, existing methods struggle to bridge the gap between LLMs' discrete representations and the continuous nature of time series forecasting.

\subsection{Multimodal Time Series Forecasting with Exogenous Text Data}
In applications, textual data can provide real-time information about events or changes that may influence time series patterns. To harness such multimodal data, several studies have begun constructing multimodal forecasting datasets \cite{xu2018stock,jia2024gpt4mts,dong2024fnspid}. For instance, \textit{From News to Forecast} \cite{wang2024news} retrieves relevant news articles based on forecasting tasks, employing LLM-powered agents to select valuable entries. Time-MMD \cite{liu2024timemmd} further expands the scope of multimodal data by providing a comprehensive dataset spanning multiple domains.

While these methods primarily focus on data collection and preprocessing stages, they have yet to fully explore deeper cross-modal fusion strategies. Although existing studies have attempted to incorporate textual information for future event forecasting~\cite{lee2025timecap,jiang2025timexl}, there remains a significant gap in understanding how to effectively integrate textual data to enhance continuous time series forecasting.

\section{Methodology}  
\subsection{Problem Formulation}  
\paragraph{Time Series Forecasting.}  
A standard time series forecasting model takes multivariate numerical data as input and outputs predictions of future values. Given a historical time series input \( X = \{X_1, \dots, X_{H}\} \in \mathbb{R}^{H \times C_{\text{in}}} \), where \( H \) denotes the length of the lookback window and \( C_{\text{in}} \) represents the number of input variables, the objective is to predict future values over a horizon \( F \). The output is denoted as \( Y = \{Y_{H+1}, \dots, Y_{H+F}\} \in \mathbb{R}^{F \times C_{\text{out}}} \), where \( C_{\text{out}} \) indicates the number of target variables. The forecasting task aims to train a model \( \mathcal{F} \) with parameters \( \theta \) that generates predictions \( \hat{Y} = \mathcal{F}_{\theta}(X) \).  

\paragraph{Multimodal Time Series Forecasting.}  
Multimodal time series forecasting incorporates exogenous textual information \( S \in \mathbb{R}^{k \times C_{\text{txt}}} \) alongside numerical inputs, where \( k \) represents the number of text sequences and \( C_{\text{txt}} \) denotes the maximum length of each sequence. For simplicity, we assume fixed-length sequences in notation. Thus, the multimodal forecasting task is formulated as \( \hat{Y} = \mathcal{F}_{\theta'}(X, S) \).

\begin{figure}[t]
    \centering
    \includegraphics[trim=10 5 10 13,clip,width=\textwidth]{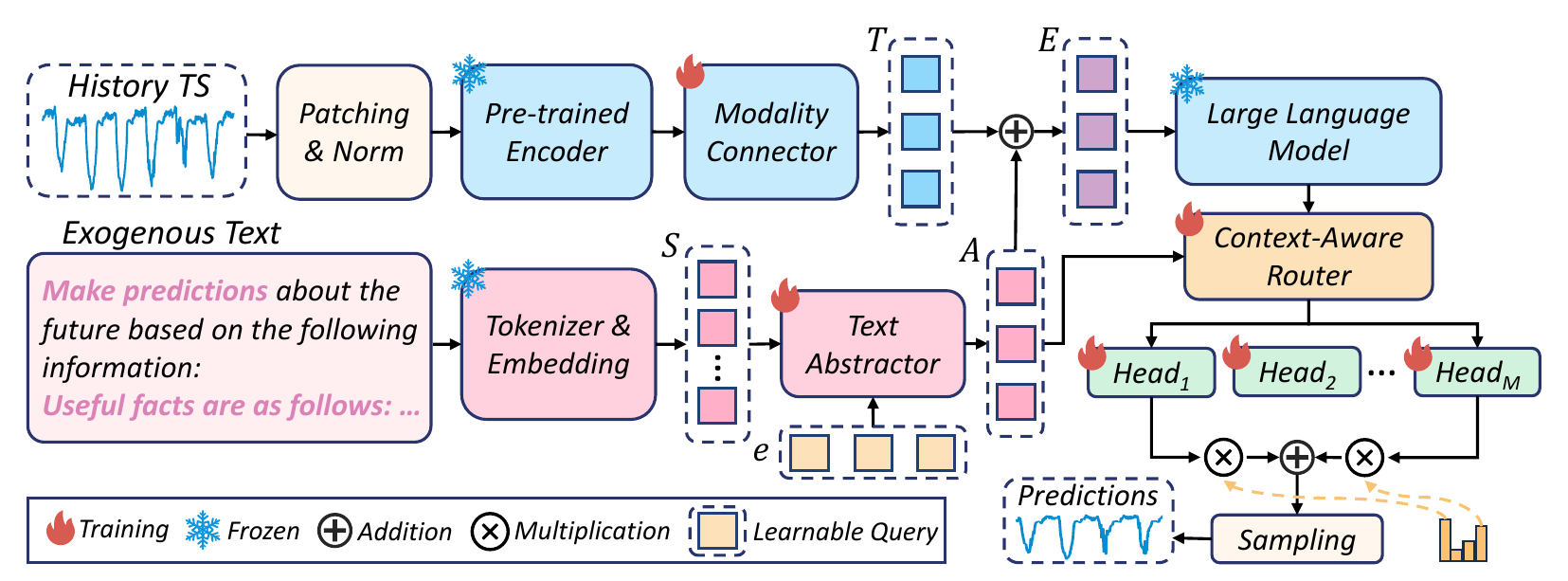}
    \caption{The CAPTime framework processes time series through patch normalization and encoding into token states, while text is embedded via LLM's tokenization. We employ frozen GPT-2~\cite{radford2019gpt2} as the LLM. Learnable queries abstract text features, which are summed with time series tokens. The unified representation undergoes LLM processing and mixture distribution modeling under negative log-likelihood optimization. During inference, forecasts are sampled from the mixed distribution.}
    \label{fig:text_example}
\end{figure}

\subsection{Time Series Token Embedding}
To enable LLMs to comprehend time series semantics, most studies~\cite{jintimellm,liu2024autotimes} employ patching techniques to segment time series data into patches treated as tokens, which are then encoded via trainable modules. However, due to the significant modality gap between the continuous nature of time series and the discrete nature of language, these simple embedding methods are insufficient for time series encoding, let alone alignment tasks. Inspired by other domains~\cite{radford2021clip}, pre-trained encoders typically project inputs into a more tractable latent space, facilitating subsequent modality alignment.

Concretely, we employ the pre-trained lightweight TSMixer~\cite{ekambaram2024ttms} as our temporal encoder, achieving pre-alignment from the temporal latent space to the text space through a modality connector $\text{MC}$. We maintain channel independence and instance normalization~\cite{nie2022time} to mitigate the complexity of pre-alignment. Given a univariate time series of length $H$, we partition it into $N_p$ non-overlapping patches of length $L_p$, resulting in the patch collection $P_H=\{X_{1:L_p},\dots,X_{H-L_p+1:H}\}\in \mathbb{R}^{N_p \times L_p}$, where $N_p=\left\lceil\frac{H}{L_p}\right\rceil
+1$. These patches are encoded as:
\begin{equation}
    T = \text{MC}(\text{TSEncoder}(P_H)),
\end{equation}
where $T\in \mathbb{R}^{N_p\times D}$ denotes the pre-aligned time series tokens, and $D$ represents the hidden dimension matching the LLM. $\text{TSEncoder}: \mathbb{R}^{L_p}\rightarrow\mathbb{R}^{D}$ refers to the pre-trained temporal encoder.

\subsection{Text Informed Token Alignment}
\paragraph{Text Abstraction.}
By combining instruction embeddings with time series tokens, the joint representation not only enables cross-modal alignment but also improves forecasting, proving simple yet effective~\cite{pan2024s2ip,suntest}. Text instructions may include dataset descriptions and statistical information~\cite{jintimellm}, or for multimodal time series forecasting, exogenous texts like news and reports~\cite{liu2024timemmd}. Existing approaches employ simple addition or concatenation of prompts with temporal tokens, enhancing encoding but lacking fine-grained alignment or detailed information incorporation. To adaptively extract relevant information for each temporal token from input prompt $S$, we introduce a text abstractor $\psi: \mathbb{R}^{N_s\times D}\rightarrow\mathbb{R}^{N_p\times D}$, where $N_s$ indicates the prompt's token count.  Specifically, we first tokenize the input prompt using the LLM to obtain $E_s\in \mathbb{R}^{N_s\times D}$. For each temporal token $T_i \in \mathbb{R}^{D}(i=1,\dots,N_p)$, we define a special \verb|<query>| token to acquire learnable embeddings $e_i\in \mathbb{R}^{D}$, which enables text abstraction $\psi$ through cross-attention:
\begin{gather}
Q_i=e_iW_{Q}, \quad K=E_sW_{K}, \quad V=E_sW_{V},\\
A_i = \text{CrossAtten}(Q_i, K, V),
\end{gather}
where $W_Q,W_K,W_V$ are learnable weight matrices with bias omitted for simplicity, and $A_i\in \mathbb{R}^{D}$ represents the contextual abstraction crucial for the $i^{th}$ time series token.

\paragraph{Text informed token embedding.}
Compared to other interactions between text and time tokens, our direct summation approach yields text-informed token embeddings $E\in \mathbb{R}^{N_p\times D}$ that preserve rich semantic information while maintaining LLM compatibility:
\begin{equation}
    E_i = T_i + A_i.
\end{equation}

\subsection{Context-Aware Probabilistic Decoding.}
Given the sequential similarity between time series and text, we feed text-informed tokens $E$ into the LLM and adopt the next-token prediction paradigm~\cite{liu2024autotimes} for forecasting subsequent time series segments. The LLM's inherent token interaction knowledge and pre-trained capability for token distribution generalization motivate our decision to freeze all LLM parameters:
\begin{equation}
    Z = \text{LLM}(E),
\end{equation}
where $Z\in \mathbb{R}^{N_p\times D}$ denotes the LLM's predicted next time series tokens.

\paragraph{Probabilistic Decoding Learning.} Existing approaches that decode LLMs' output tokens into time series predictions typically rely on exact value mappings~\cite{zhou2023gpt4ts,jintimellm}, which prove difficult to learn. However, benefiting from LLMs' inherent probabilistic forecasting capabilities and training objectives, models demonstrate superior performance in learning distribution mappings compared to exact values. We therefore formulate decoding as a learnable mapping $\Phi_{\theta}$ from token space to per-timestep distributions (e.g., Student's t-distribution) to capture predictive uncertainty:
\begin{equation}
    Z_i \xrightarrow{\Phi_{\theta}} \prod_{j=1}^{L_p} p(\hat{Y}_{i,j} \mid \mu_{i,j}, \sigma_{i,j}, \nu_{i,j}), 
\end{equation}
where $\hat{Y}_{i,j}$ indicates the forecasts at the $j^{th}$ timestep within the next patch from the $i^{th}$ token $Z_i$, with $\mu_{i,j}$, $\sigma_{i,j}$, and $\nu_{i,j}$ parameterizing the distribution's location, scale, and degrees of freedom.

\paragraph{Context-Aware Mixture Distribution Modeling.}
To handle the complex dynamics of time series, we design $K$ distinct distribution decoders and employ mixture of experts \cite{fedus2022switch} for generalized learning. The extracted textual information $A$ controls a gating module $\text{Gate}$ that adaptively assigns distribution decoders to different tokens. This context-aware approach outperforms unimodal decoding through enhanced semantic awareness, effectively capturing exogenous variations. Our sparse routing structure selects the top-$K$ experts among $M$ candidates for each token:
\begin{gather}
h_i=[\mu_{i}, \sigma_{i}, \nu_{i}]=\sum_{m=1}^M \left( g_{i,m} \, \text{Head}_m\left(Z_i\right) \right),\\
g_{i,m} = \begin{cases}
    s_{i,m}, & s_{i,m} \in \text{Topk}(\{s_{i,j} \mid 1\leq j \leq M\}, K), \\
    0, & \text{otherwise},
\end{cases}\\
s_i = \text{Softmax}_i(\text{Gate}(A_i)),
\end{gather}
where $h_i\in \mathbb{R}^{3L_p}$ comprises the predicted distribution parameters. The expert weights $g_{i,m}$ derive from the normalized gating outputs $s_i\in \mathbb{R}^{3L_p}$ via \text{Topk} selection, and each $\textup{Head}_m$ implements the decoding mapping $\Phi_{\theta}$. We train this distribution decoder using negative log-likelihood minimization. To promote expert diversity and balanced utilization~\cite{fedus2022switch}, we incorporate a load balancing loss $\mathcal{L}_b$, yielding the following loss:
\begin{equation}
    \mathcal{L} = -\sum_{i=1}^{N_p} log(\prod_{j=1}^{L_p} p(\hat{Y}_{i,j}\mid \mu_{i,j}, \sigma_{i,j}, \nu_{i,j}))+\mathcal{L}_b.
\end{equation}
During inference, we select the value with the maximum probability from the fused distribution as the prediction. Owing to the next-token prediction paradigm, our model requires only a single training procedure across varying prediction horizons, with autoregressive generation producing predictions until the target horizon is reached.

\section{Experiments}
We conduct extensive time series forecasting experiments, including multimodal forecasting (Sec. \ref{Multimodal Forecasting}), short-term forecasting (Sec. \ref{Short-term Forecasting}), long-term forecasting (Sec. \ref{Long-term Forecasting}), few-shot forecasting (Sec. \ref{Few-shot Forecasting}), and zero-shot forecasting (Sec. \ref{Zero-shot Forecasting}), to validate the robustness and superior performance of our method. Through comprehensive ablation studies (Sec. \ref{Ablation Study}), we demonstrate the effectiveness of our proposed modules and technical implementations. Additionally, method analysis (Sec. \ref{Method Analysis}) illustrates how textual integration enhances multimodal forecasting capabilities.

\paragraph{Baselines.} We compare CAPTime with several state-of-the-art approaches: LLM-based methods, including S\textsuperscript{2}IP-LLM~\cite{pan2024s2ip}, TimeCMA~\cite{liu2024timecma}, AutoTimes~\cite{liu2024autotimes}, Time-LLM~\cite{jintimellm}, and GPT4TS~\cite{zhou2023gpt4ts}; and deep forecasters, including iTransformer~\cite{liuitransformer}, DLinear~\cite{zeng2023dlinear}, PatchTST~\cite{nie2022time}, TimesNet~\cite{wutimesnet}, and FEDformer~\cite{zhou2022fedformer}. For short-term forecasting tasks, we additionally include N-HiTS~\cite{challu2023nhits} as a strong baseline.

\subsection{Multimodal Forecasting}
\label{Multimodal Forecasting}

\paragraph{Setup.} To evaluate CAPTime's multimodal forecasting efficacy, we conduct experiments on seven datasets with exogenous texts~\cite{liu2024timemmd}: Agriculture, Economy, Energy, Health, Security, SocialGood, and TrafficText. Following Time-MMD~\cite{liu2024timemmd}, we set the input sequence length to 8, except for the Health and Energy datasets, which use 36. The evaluation horizons include four scales: \{12, 24, 36, 48\} for Health and Energy, and \{6, 8, 10, 12\} for other datasets. Technically, we adhere to Time-MMD's evaluation framework for fair comparison, using MSE and MAE metrics. Notably, benefiting from the next-token prediction paradigm, both CAPTime and AutoTimes~\cite{liu2024autotimes} require only one single training per dataset for all horizons, while other baselines require retraining for each horizon.

\paragraph{Results.} Experimental results are presented in Tab. \ref{tab:text_performance}. CAPTime outperforms all baselines across datasets, demonstrating superior multimodal forecasting capabilities. Specifically, CAPTime achieves an average 19.4\% improvement in MSE over the second-best method, FEDformer~\cite{zhou2022fedformer}. We observe that existing LLM-based methods show no significant advantage over deep forecasters, indicating insufficient exploitation of exogenous textual information. In contrast, CAPTime's carefully designed text abstraction mechanism and context-aware forecasting demonstrate enhanced textual understanding and superior multimodal forecasting performance.

\vspace{-0.3cm}
\begin{table}[htbp]
\centering
\caption{Multimodal forecasting performance comparison. \textbf{Bold}: best, \underline{underline}: second best.}
\fontsize{27pt}{31pt}\selectfont
\label{tab:text_performance}
\renewcommand{\arraystretch}{1.2}
\setlength{\arrayrulewidth}{1.7pt}  
\setlength{\heavyrulewidth}{3.5pt}  
\setlength{\lightrulewidth}{1.7pt}
\setlength{\cmidrulewidth}{1.7pt}
\resizebox{1.0\textwidth}{!}{
\begin{tabular}{c|cc|cc|cc|cc|cc|cc|cc|cc|cc|cc}
\toprule
\multicolumn{1}{c|}{Models} & \multicolumn{2}{c|}{CAPTime} & \multicolumn{2}{c|}{S\textsuperscript{2}IP-LLM} & \multicolumn{2}{c|}{AutoTimes} & \multicolumn{2}{c|}{Time-LLM} & \multicolumn{2}{c|}{GPT4TS} & \multicolumn{2}{c|}{iTransformer} & \multicolumn{2}{c|}{DLinear} & \multicolumn{2}{c|}{PatchTST} & \multicolumn{2}{c|}{TimesNet} & \multicolumn{2}{c}{FEDformer} \\
\cmidrule(lr){1-21}
\multicolumn{1}{c|}{Metric} & MSE & MAE & MSE & MAE & MSE & MAE & MSE & MAE & MSE & MAE & MSE & MAE & MSE & MAE & MSE & MAE & MSE & MAE & MSE & MAE \\
\midrule
\multirow{1}{*}{{Agri.}} 
& \textbf{0.193} & \textbf{0.291} & 0.333 & 0.470 & 0.328 & 0.472 & 0.309 & 0.456 & 0.342 & 0.485 & 0.317 & 0.464 & 0.400 & 0.528 & 0.328 & 0.475 & 0.292 & 0.439 & \underline{0.219} & \underline{0.357} \\
\multirow{1}{*}{{Econ.}} 
& \textbf{0.121} & \textbf{0.217} & 0.389 & 0.573 & 0.384 & 0.571 & 0.379 & 0.570 & 0.384 & 0.572 & 0.335 & 0.544 & 0.418 & 0.609 & 0.353 & 0.557 & 0.335 & 0.541 & \underline{0.245} & \underline{0.431} \\
\multirow{1}{*}{{Ener.}} 
& \textbf{0.261} & \textbf{0.369} & 0.356 & 0.452 & 0.330 & 0.434 & 0.318 & 0.427 & 0.316 & 0.422 & 0.301 & 0.419 & 0.383 & 0.451 & 0.281 & 0.404 & 0.325 & 0.438 & \underline{0.262} & \underline{0.377} \\
\multirow{1}{*}{{Heal.}} 
& \textbf{1.154} & \textbf{0.719} & 1.387 & 0.998 & 1.420 & 0.870 & 1.479 & 0.816 & 1.395 & 0.797 & 1.619 & 0.798 & 1.774 & 0.880 & 1.486 & \underline{0.776} & 1.749 & 0.833 & \underline{1.377} & 0.844 \\
\multirow{1}{*}{{Secu.}} 
& \textbf{70.33} & \textbf{4.008} & 126.8 & 5.780 & 116.8 & 5.128 & 108.6 & 4.854 & \underline{107.6} & \underline{4.746} & 115.8 & 5.444 & 108.8 & 5.075 & 112.1 & 5.159 & 129.5 & 5.989 & 115.9 & 5.243 \\
\multirow{1}{*}{{Soci.}} 
& \textbf{0.838} & \textbf{0.377} & 1.078 & 0.580 & 0.963 & 0.530 & 0.914 & \underline{0.490} & \underline{0.871} & 0.500 & 1.117 & 0.501 & 0.956 & 0.527 & 1.032 & 0.506 & 1.243 & 0.545 & 0.882 & 0.491 \\
\multirow{1}{*}{{Traf.}} 
& \textbf{0.194} & \textbf{0.291} & 0.311 & 0.446 & 0.277 & 0.414 & 0.236 & 0.368 & 0.283 & 0.429 & 0.221 & 0.343 & 0.361 & 0.502 & 0.224 & 0.354 & \underline{0.207} & \underline{0.326} & 0.221 & 0.368 \\
\bottomrule
\end{tabular}
}
\end{table}

\subsection{Short-term Forecasting}
\label{Short-term Forecasting}
\paragraph{Setup.} We evaluate short-term forecasting performance on the M4 dataset~\cite{makridakis2020m4}, a standard benchmark comprising tasks across multiple frequencies. Following established protocols~\cite{wutimesnet}, we assess forecasting horizons ranging from 6 to 48 steps, with input sequences set to twice the horizon length. Evaluation metrics include symmetric mean absolute percentage error (SMAPE), mean absolute scaled error (MASE), and overall weighted average (OWA).
\paragraph{Results.} As shown in Tab. \ref{tab:short_performance}, CAPTime outperforms all baseline methods in short-term forecasting tasks. These results highlight the model's effectiveness in capturing temporal patterns in unimodal settings, complementing the multimodal capabilities, owing to its advanced distribution modeling approach.

\vspace{-0.2cm}
\begin{table}[htbp]
\centering
\caption{Short-term forecasting performance comparison. \textbf{Bold}: best, \underline{underline}: second best.}
\fontsize{24pt}{28pt}\selectfont
\label{tab:short_performance}
\renewcommand{\arraystretch}{1.2}
\setlength{\arrayrulewidth}{1.4pt}  
\setlength{\heavyrulewidth}{3.2pt}  
\setlength{\lightrulewidth}{1.4pt}
\setlength{\cmidrulewidth}{1.4pt}
\resizebox{1.0\textwidth}{!}{
\begin{tabular}{c|c|c|c|c|c|c|c|c|c|c|c|c}
\toprule
\multicolumn{2}{c|}{Models} & CAPTime & S\textsuperscript{2}IP-LLM  & AutoTimes & Time-LLM & GPT4TS & iTransformer & DLinear & PatchTST & N-HiTS & TimesNet & FEDformer   \\
\cmidrule(lr){1-13}
\multirow{3}{*}{\rotatebox[origin=c]{90}{Average}} 
& SMAPE & \textbf{11.832} & 12.021 & \underline{11.865} & 12.494 & 12.690 & 12.142 & 13.639 & 12.059 & 12.035 & 12.880 & 13.160 \\
& MASE & \textbf{1.580} & 1.612 & \underline{1.591} & 1.731 & 1.808 & 1.631 & 2.095 & 1.623 & 1.625 & 1.836 & 1.775 \\
& OWA & \textbf{0.850} & 0.857 & \underline{0.853} & 0.913 & 0.940 & 0.874 & 1.051 & 0.869 & 0.869 & 0.955 & 0.949 \\
\bottomrule
\end{tabular}
}
\end{table}

\subsection{Long-term Forecasting}
\label{Long-term Forecasting}
\paragraph{Setup.} To assess CAPTime's long-term forecasting performance, we conduct experiments on seven standard datasets~\cite{wu2021autoformer}: ETTh1, ETTh2, ETTm1, ETTm2, Weather, Electricity, and ILI. We compare against multiple state-of-the-art methods using forecasting horizons of \{24, 36, 48, 60\} for ILI and \{96, 192, 336, 720\} for other datasets.
\paragraph{Results.} Tab. \ref{tab:long_performance} presents the experimental results, where CAPTime achieves consistently strong performance across most datasets. The model shows 4.7\% and 5.2\% improvements in MAE over PatchTST~\cite{nie2022time} and S\textsuperscript{2}IP-LLM~\cite{pan2024s2ip}, respectively. Notably, compared to the autoregressive AutoTimes~\cite{liu2024autotimes}, CAPTime achieves an 8.5\% performance gain. These improvements stem from our pre-trained temporal encoder's effective modeling of temporal dependencies and its ability to model future time series distributions.

\vspace{-0.2cm}
\begin{table}[htbp]
\centering
\caption{Long-term forecasting performance comparison. \textbf{Bold}: best, \underline{underline}: second best.}
\fontsize{27pt}{31pt}\selectfont
\label{tab:long_performance}
\renewcommand{\arraystretch}{1.2}
\setlength{\arrayrulewidth}{1.7pt}
\setlength{\heavyrulewidth}{3.5pt}
\setlength{\lightrulewidth}{1.7pt}
\setlength{\cmidrulewidth}{1.7pt}
\resizebox{1.0\textwidth}{!}{
\begin{tabular}{c|cc|cc|cc|cc|cc|cc|cc|cc|cc|cc}
\toprule
\multicolumn{1}{c|}{Models} & \multicolumn{2}{c|}{CAPTime} & \multicolumn{2}{c|}{S\textsuperscript{2}IP-LLM} & \multicolumn{2}{c|}{TimeCMA} & \multicolumn{2}{c|}{AutoTimes} & \multicolumn{2}{c|}{Time-LLM} & \multicolumn{2}{c|}{GPT4TS} & \multicolumn{2}{c|}{iTransformer} & \multicolumn{2}{c|}{DLinear} & \multicolumn{2}{c|}{PatchTST} & \multicolumn{2}{c}{TimesNet}\\
\cmidrule(lr){1-21}
\multicolumn{1}{c|}{Metric} & MSE & MAE & MSE & MAE & MSE & MAE & MSE & MAE & MSE & MAE & MSE & MAE & MSE & MAE & MSE & MAE & MSE & MAE & MSE & MAE \\
\midrule
\multirow{1}{*}{{ETTh1}}
& \textbf{0.396} & \textbf{0.419} & 0.418 & 0.436 & 0.422 & 0.445 & 0.427 & 0.439 & 0.426 & 0.435 & 0.427 & \underline{0.426} & 0.451 & 0.462 & 0.422 & 0.437 & \underline{0.413} & 0.430 & 0.458 & 0.450 \\
\multirow{1}{*}{ETTh2} 
& \textbf{0.341} & \textbf{0.381} & 0.355 & 0.399 & \underline{0.352} & 0.405 & 0.358 & 0.401 & 0.361 & 0.398 & 0.354 & \underline{0.394} & 0.382 & 0.414 & 0.431 & 0.446 & 0.381 & 0.411 & 0.414 & 0.427\\
\multirow{1}{*}{ETTm1}
& \textbf{0.345} & \textbf{0.377} & \underline{0.346} & 0.382 & 0.365 & 0.398 & 0.376 & 0.405 & 0.354 & 0.384 & 0.352 & 0.383 & 0.370 & 0.399 & 0.357 & 0.378 & 0.351 & \underline{0.380} & 0.400 & 0.406\\
\multirow{1}{*}{ETTm2} 
& \textbf{0.257} & \textbf{0.304} & \underline{0.262} & 0.326 & 0.293 & 0.346 & 0.278 & 0.331 & 0.275 & 0.334 & 0.266 & 0.326 & 0.272 & 0.331 & 0.267 & 0.333 & 0.267 & \underline{0.325} & 0.291 & 0.333\\
\multirow{1}{*}{Weather} 
& \underline{0.229} & \textbf{0.251} & \textbf{0.228} & \underline{0.265} & 0.232 & 0.275 & 0.246 & 0.284 & 0.237 & 0.269 & 0.237 & 0.270 & 0.304 & 0.335 & 0.248 & 0.300 & 0.230 & \underline{0.265} & 0.259 & 0.287\\
\multirow{1}{*}{ECL} 
& 0.185 & 0.269 & \underline{0.166} & \underline{0.262} & 0.190 & 0.295 & 0.175 & 0.270 & 0.167 & 0.264 & 0.167 & 0.263 & 0.203 & 0.298 & \underline{0.166} & 0.263 & \textbf{0.161} & \textbf{0.252} & 0.192 & 0.295\\
\multirow{1}{*}{{ILI}}
& \underline{2.003} & \textbf{0.835} & 2.332 & 1.005 & \textbf{1.928} & \underline{0.917} & 2.723 & 1.082 & 2.432 & 1.012 & 2.623 & 1.060 & 2.444 & 1.203 & 2.616 & 1.090 & 2.388 & 1.011 & 2.139 & 0.931\\
\bottomrule
\end{tabular}
}
\end{table}
\vspace{-0.2cm}

\subsection{Few-shot Forecasting}
\label{Few-shot Forecasting}
\paragraph{Setup.} Leveraging LLMs' extensive pretrained knowledge and generalization ability, LLM-based approaches exhibit robust performance under data scarcity. To evaluate CAPTime's enhancement of LLMs' distribution modeling capacity, we perform few-shot forecasting experiments on both unimodal (ETTh1, ETTh2) and multimodal (Energy/Health/TrafficText) benchmarks. We utilize 5\% of training data for ETTh1 and ETTh2, and 10\% for other smaller datasets.
\paragraph{Results.} As shown in Tab. \ref{tab:few_performance}, CAPTime outperforms state-of-the-art baselines across all evaluation scenarios. For unimodal forecasting, it achieves MSE reductions of 11.7\% and 9.8\% over Time-LLM~\cite{jintimellm} and S\textsuperscript{2}IP-LLM~\cite{pan2024s2ip}, respectively. In multimodal settings, CAPTime shows substantial gains over GPT4TS and deep learning models such as PatchTST~\cite{nie2022time}. These results further demonstrate our method's effectiveness in extracting key information from exogenous texts and its context-aware distribution modeling, particularly in few-shot multimodal forecasting.
\vspace{-0.2cm}
\begin{table}[htbp]
\centering
\caption{Few-shot forecasting performance comparison. \textbf{Bold}: best, \underline{underline}: second best.}
\fontsize{25pt}{31pt}\selectfont
\label{tab:few_performance}
\renewcommand{\arraystretch}{1.2}
\setlength{\arrayrulewidth}{1.7pt}
\setlength{\heavyrulewidth}{3.5pt}
\setlength{\lightrulewidth}{1.7pt}
\setlength{\cmidrulewidth}{1.7pt}
\resizebox{1.0\textwidth}{!}{
\begin{tabular}{c|*{9}{cc|}cc}
\toprule
\multicolumn{1}{c|}{Models} & \multicolumn{2}{c|}{CAPTime} & \multicolumn{2}{c|}{S\textsuperscript{2}IP-LLM} & \multicolumn{2}{c|}{AutoTimes} & \multicolumn{2}{c|}{Time-LLM} & \multicolumn{2}{c|}{GPT4TS} & \multicolumn{2}{c|}{iTransformer} & \multicolumn{2}{c|}{DLinear} & \multicolumn{2}{c|}{PatchTST} & \multicolumn{2}{c|}{TimesNet} & \multicolumn{2}{c}{FEDformer} \\

\cmidrule(lr){1-21}

\multicolumn{1}{c|}{Metric} & MSE & MAE & MSE & MAE & MSE & MAE & MSE & MAE & MSE & MAE & MSE & MAE & MSE & MAE & MSE & MAE & MSE & MAE & MSE & MAE \\
\midrule
\multirow{1}{*}{{ETTh1}} 
& \textbf{0.599} & \textbf{0.505} & 0.650 & 0.550 & 0.707 & 0.592 & \underline{0.648} & \underline{0.549} & 0.681 & 0.560 & 1.070 & 0.710 & 0.750 & 0.611 & 0.695 & 0.569 & 0.925 & 0.647 & 0.658 & 0.562 \\
\multirow{1}{*}{{ETTh2}} 
& \textbf{0.335} & \textbf{0.379} & \underline{0.380} & \underline{0.413} & 0.420 & 0.436 & 0.398 & 0.426 & 0.400 & 0.433 & 0.488 & 0.475 & 0.827 & 0.615 & 0.439 & 0.448 & 0.463 & 0.454 & 0.463 & 0.454 \\
\multirow{1}{*}{{Ener.}} 
& \textbf{0.386} & \textbf{0.474} & 0.687 & 0.639 & 0.470 & 0.516 & 0.589 & 0.583 & 0.478 & 0.519 & 0.462 & 0.513 & \underline{0.432} & \underline{0.487} & 0.561 & 0.570 & 0.518 & 0.573 & 0.483 & 0.519 \\
\multirow{1}{*}{{Heal.}} 
& \textbf{1.389} & \textbf{0.813} & 2.655 & 1.178 & 2.415 & 1.128 & 2.113 & 1.036 & 2.143 & 1.034 & 1.663 & \underline{0.868} & 2.209 & 1.083 & \underline{1.621} & 0.870 & 2.242 & 1.052 & 2.174 & 1.078 \\
\multirow{1}{*}{{Traff.}} 
& \textbf{0.314} & \textbf{0.431} & 0.627 & 0.615 & 0.466 & 0.592 & 0.620 & 0.641 & \underline{0.406} & \underline{0.529} & 0.412 & 0.530 & 0.504 & 0.589 & 0.425 & 0.541 & 0.463 & 0.561 & 0.755 & 0.718 \\
\bottomrule
\end{tabular}
}
\end{table}

\subsection{Zero-shot Forecasting}
\label{Zero-shot Forecasting}
\paragraph{Setup.} Beyond few-shot scenarios, LLMs demonstrate strong zero-shot generation capabilities. However, existing methods have not fully exploited LLMs' generalization potential, resulting in inferior zero-shot performance compared to deep forecasters. To evaluate CAPTime's enhanced generalization, we adopt Time-LLM's experimental setup~\cite{jintimellm} on multiple ETT datasets~\cite{wu2021autoformer} and multimodal datasets (Health, Energy). The evaluation involves training on a source dataset and testing on target datasets without any training samples to assess zero-shot performance.

\paragraph{Results.} Results in Tab. \ref{tab:zero_performance} demonstrate CAPTime's superior generalization capability. Our method outperforms all baselines across experimental settings, particularly in multimodal zero-shot scenarios, achieving 18.2\% and 17.8\% improvements over S\textsuperscript{2}IP-LLM~\cite{pan2024s2ip} and AutoTimes~\cite{liu2024autotimes}. Notably, CAPTime exhibits excellent generalization in both data-scarce zero-shot and few-shot settings, surpassing the second-best method, PatchTST~\cite{nie2022time} by 21.8\% in few-shot and 11.1\% in zero-shot scenarios. These findings confirm the efficacy of our proposed text abstraction for token alignment and distribution modeling, while enhancing the model's robustness in data-scarce environments.

\vspace{-0.2cm}
\begin{table}[htbp]
\centering
\caption{Zero-shot forecasting performance comparison. \textbf{Bold}: best, \underline{underline}: second best.}
\fontsize{32pt}{39pt}\selectfont
\label{tab:zero_performance}
\renewcommand{\arraystretch}{1.2}
\setlength{\arrayrulewidth}{2.0pt}  
\setlength{\heavyrulewidth}{4.0pt}  
\setlength{\lightrulewidth}{2pt}
\setlength{\cmidrulewidth}{2pt}
\resizebox{1.0\textwidth}{!}{
\begin{tabular}{c|cc|cc|cc|cc|cc|cc|cc|cc|cc}
\toprule
\multicolumn{1}{c|}{Models} & \multicolumn{2}{c|}{CAPTime} & \multicolumn{2}{c|}{S\textsuperscript{2}IP-LLM} & \multicolumn{2}{c|}{AutoTimes} & \multicolumn{2}{c|}{Time-LLM} & \multicolumn{2}{c|}{GPT4TS} & \multicolumn{2}{c|}{iTransformer} & \multicolumn{2}{c|}{DLinear} & \multicolumn{2}{c|}{PatchTST} & \multicolumn{2}{c}{TimesNet}   \\

\cmidrule(lr){1-19}

\multicolumn{1}{c|}{Metric} & MSE & MAE & MSE & MAE & MSE & MAE & MSE & MAE & MSE & MAE & MSE & MAE & MSE & MAE & MSE & MAE & MSE & MAE\\
\midrule
\multirow{1}{*}{{ETTh1$\to$ ETTh2}} 
& \textbf{0.358} & \textbf{0.392} & 0.403 & 0.417 & \underline{0.368} & \underline{0.402} & 0.384 & 0.409 & 0.406 & 0.422 & 0.457 & 0.455 & 0.493 & 0.488 & 0.380 & 0.405 & 0.421 & 0.431  \\
\multirow{1}{*}{{ETTh2$\to$ETTh1}} 
& \textbf{0.449} & \textbf{0.451} & 0.669 & 0.560 & 0.789 & 0.623 & 0.663 & 0.540 & 0.757 & 0.578 & 0.868 & 0.625 & 0.703 & 0.574 & \underline{0.565} & \underline{0.513} & 0.865 & 0.621  \\
\multirow{1}{*}{{ETTh1$\to$ETTm2}} 
& \textbf{0.293} & \textbf{0.343} & 0.325 & \underline{0.360} & 0.336 & 0.381 & 0.317 & 0.370 & 0.325 & 0.363 & 0.360 & 0.390 & 0.415 & 0.452 & \underline{0.314} & \underline{0.360} & 0.327 & 0.361  \\
\multirow{1}{*}{{ETTh2$\to$ETTm2}} 
& \textbf{0.294} & \textbf{0.343} & 0.327 & \underline{0.363} & \underline{0.322} & 0.370 & 0.339 & 0.371 & 0.335 & 0.370 & 0.335 & 0.382 & 0.328 & 0.386 & 0.325 & 0.365 & 0.342 & 0.376  \\
\multirow{1}{*}{{ETTm1$\to$ETTm2}} 
& \textbf{0.267} & \textbf{0.319} & 0.304 & 0.347 & 0.304 & 0.350 & 0.311 & 0.343 & 0.313 & 0.348 & 0.319 & 0.363 & 0.335 & 0.389 & \underline{0.296} & \underline{0.334} & 0.322 & 0.354  \\
\multirow{1}{*}{{ETTm2$\to$ETTm1}} 
& \textbf{0.489} & \textbf{0.451} & 0.622 & 0.532 & 0.597 & 0.509 & 0.588 & 0.503 & 0.769 & 0.567 & 0.706 & 0.572 & 0.649 & 0.537 & \underline{0.568} & \underline{0.492} & 0.769 & 0.567  \\
\multirow{1}{*}{{Health$\to$Energy}} 
& \textbf{0.383} & \textbf{0.467} & 0.478 & 0.517 & 0.531 & 0.553 & \underline{0.426} & \underline{0.482} & 0.478 & 0.521 & 0.468 & 0.511 & 0.518 & 0.532 & 0.447 & 0.519 & 0.476 & 0.540  \\
\multirow{1}{*}{{Energy$\to$Health}} 
& \textbf{1.560} & \textbf{0.868} & 2.120 & 1.052 & 1.874 & 0.980 & 1.894 & 0.999 & 1.925 & 1.022 & 1.917 & 1.024 & 2.166 & 1.079 & \underline{1.699} & \underline{0.939} & 1.986 & 1.057  \\
\bottomrule
\end{tabular}
}
\end{table}

\subsection{Ablation Study}
\label{Ablation Study}
We conduct comprehensive ablation studies to analyze CAPTime's effectiveness, with results shown in Tab. \ref{tab:ablation}. The evaluation covers two long-term forecasting and two multimodal forecasting datasets to verify our model design and LLM utilization strategy.

\vspace{-0.2cm}
\begin{table}[htbp]
\centering
\caption{Ablation study results on four datasets. \textbf{Bold}: the best.}
\fontsize{7pt}{7pt}\selectfont
\label{tab:ablation}
{
\begin{tabular}{c|cc|cc|cc|cc}
\toprule
\multicolumn{1}{c|}{Datasets} & \multicolumn{2}{c|}{ETTm1} & \multicolumn{2}{c|}{Weather} & \multicolumn{2}{c|}{Health} & \multicolumn{2}{c}{Energy} \\
\cmidrule(lr){1-9}
\multicolumn{1}{c|}{Metric} & MSE & MAE & MSE & MAE & MSE & MAE & MSE & MAE \\
\midrule
\textbf{A.1} w/o TSEncoder          & 0.354 & 0.377 & 0.246 & 0.267 & 1.277 & 0.770 & 0.274 & 0.376 \\
\textbf{A.2} w/o Text Abstraction   & 0.356 & 0.378 & 0.231 & 0.252 & 1.326 & 0.809 & 0.324 & 0.425 \\
\textbf{A.3} w/o Context Aware      & 0.354 & 0.377 & 0.230 & 0.253 & 1.322 & 0.807 & 0.316 & 0.404 \\
\textbf{A.4} w/ Point Forecasting    & 0.357 & 0.385 & 0.233 & 0.266 & 1.334 & 0.780 & 0.288 & 0.393 \\
\midrule
\textbf{B.1} Finetunable LLM        & 0.359 & 0.381 & 0.239 & 0.260 & 1.191 & 0.723 & 0.276 & 0.384 \\
\textbf{B.2} Random Initialization        & 0.355 & 0.380 & 0.234 & 0.256 & 1.290 & 0.767 & 0.284 & 0.387 \\
\textbf{B.3} w/o LLM                & 0.357 & 0.377 & 0.248 & 0.271 & 1.233 & 0.738 & 0.281 & 0.388 \\
\textbf{B.4} LLM2Attn               & 0.348 & 0.379 & 0.254 & 0.277 & 1.221 & 0.726 & 0.278 & 0.381 \\
\midrule
CAPTime                 & \textbf{0.345} & \textbf{0.377} & \textbf{0.229} & \textbf{0.251} & \textbf{1.154} & \textbf{0.719} & \textbf{0.261} & \textbf{0.369}
 \\
\bottomrule
\end{tabular}
}
\end{table}

\paragraph{Validation of Model Design.} We analyze four model variants (\textbf{A.1}--\textbf{A.4}): 
\textbf{A.1} replaces the pre-trained temporal encoder with a trainable multilayer perceptron, showing performance degradation that confirms the importance of pre-trained time series knowledge. 
\textbf{A.2} omits the text abstraction module, feeding temporal tokens to the LLM, which leads to a significant performance decline, particularly. 
\textbf{A.3} excludes text abstraction from the gating mechanism in distribution modeling, also causing a notable performance drop. Ablation studies in \textbf{A.2} and \textbf{A.3} demonstrate that our context-aware distribution modeling with text abstraction markedly enhances cross-modal semantic fusion. 
\textbf{A.4} employs conventional point forecasting, yielding inferior results that underscore the superiority of probabilistic distribution forecasting for temporal modeling.

\paragraph{Validation of LLM Utilization.} Following prior work on LLM effectiveness~\cite{tan2024llmuseful}, we evaluate four configurations: 
\textbf{B.1} sets the LLM backbone's parameters as trainable; 
\textbf{B.2} uses randomly initialized weights instead of pre-trained knowledge; 
\textbf{B.3} removes the LLM entirely, decoding text-informed temporal tokens directly; 
\textbf{B.4} replaces the LLM with a simple trainable attention layer. 
Results confirm that CAPTime effectively leverages LLM knowledge without requiring additional fine-tuning, which is often ineffective with limited data. Compared to non-autoregressive and point forecasting approaches, our autoregressive framework with multi-distribution forecasting better harnesses pre-trained LLM knowledge. Similar to distribution modeling in language tasks, our approach more effectively unlocks the LLM's potential for temporal forecasting, particularly in multimodal scenarios.

\subsection{Method Analysis}
\label{Method Analysis}

\begin{figure}[htbp]
    \centering
    \begin{subfigure}[b]{0.28\textwidth}  
        \centering
        \includegraphics[trim=20 20 20 20,clip,width=\textwidth]{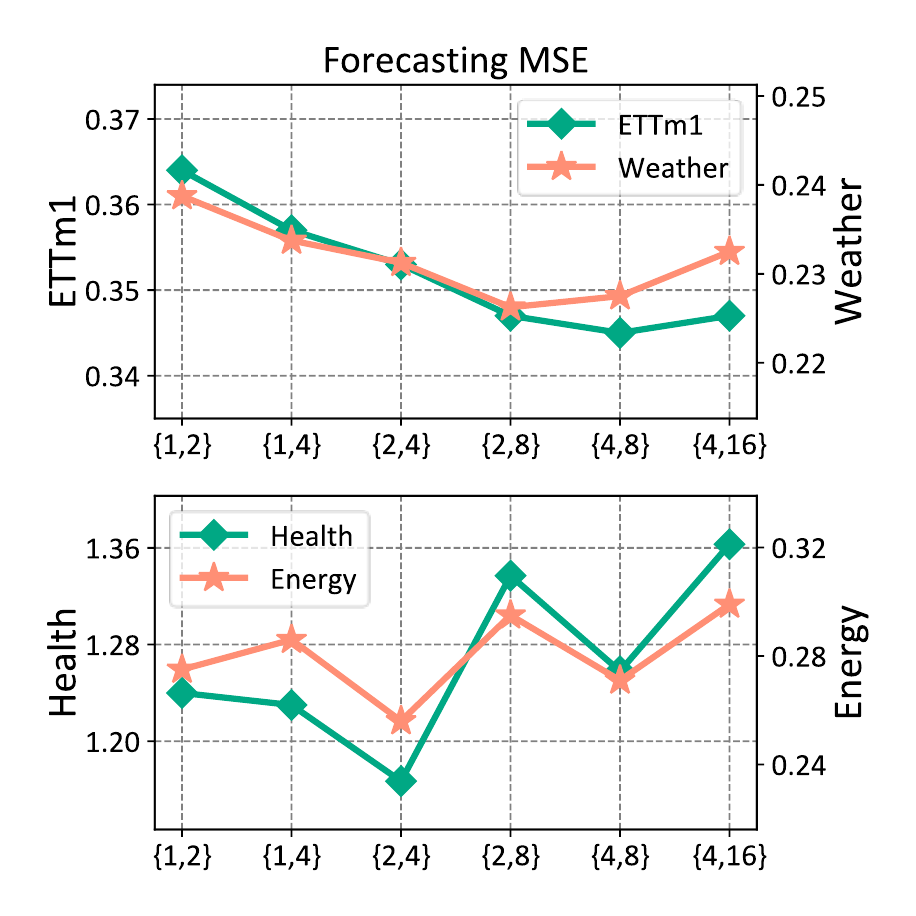}
        \caption{Performance analysis of hyper-parameter settings.}
        \label{sub_hyper}
    \end{subfigure}
    \hspace{0.1cm}
    \begin{subfigure}[b]{0.30\textwidth}  
        \centering
        \includegraphics[trim=33 30 27 30,clip,width=\textwidth]{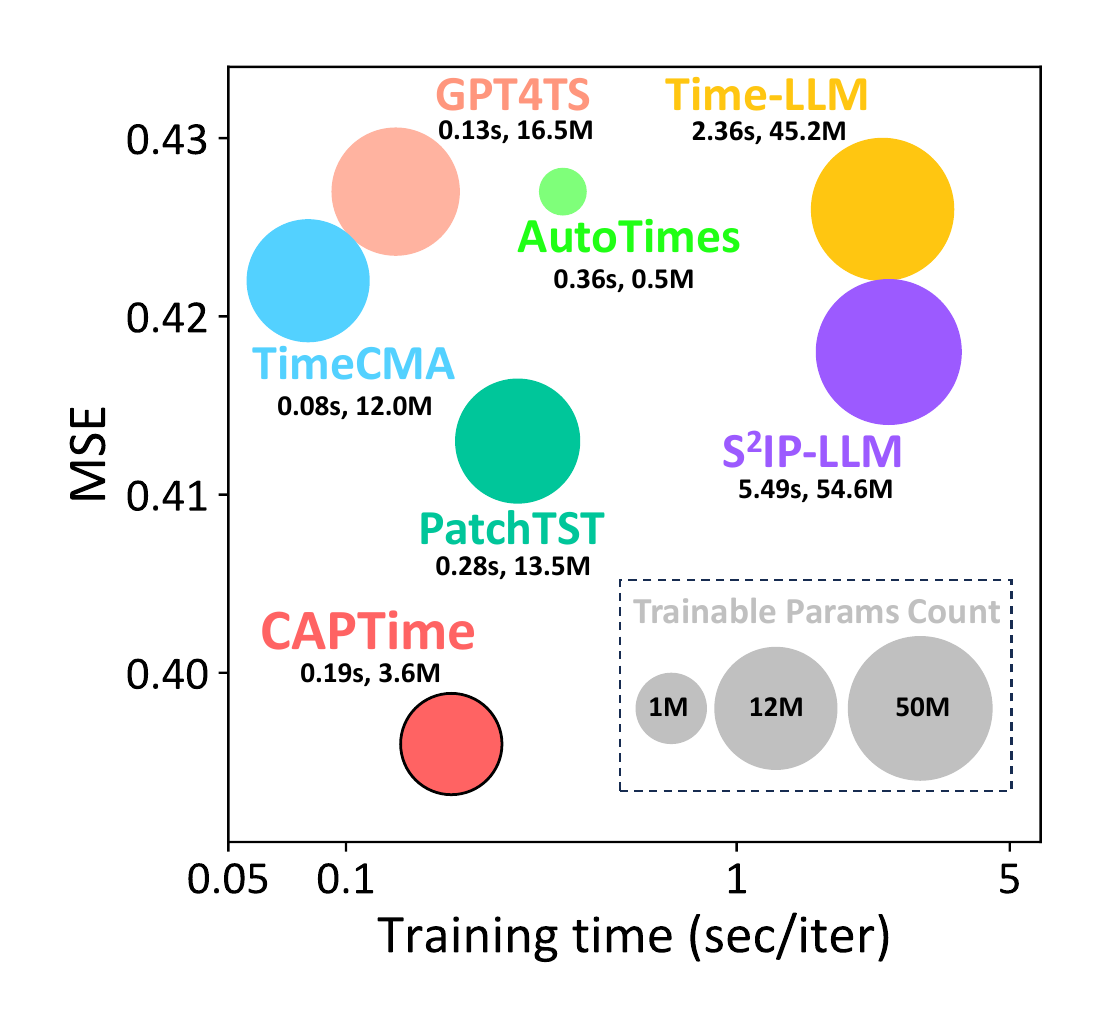}
        \caption{Computational efficiency evaluation on ETTh1.}
        \label{sub_efficiency}
    \end{subfigure}
    \hspace{0.1cm}
    \begin{subfigure}[b]{0.36\textwidth}  
        \centering
        \includegraphics[trim=20 25 20 22,clip,width=\textwidth]{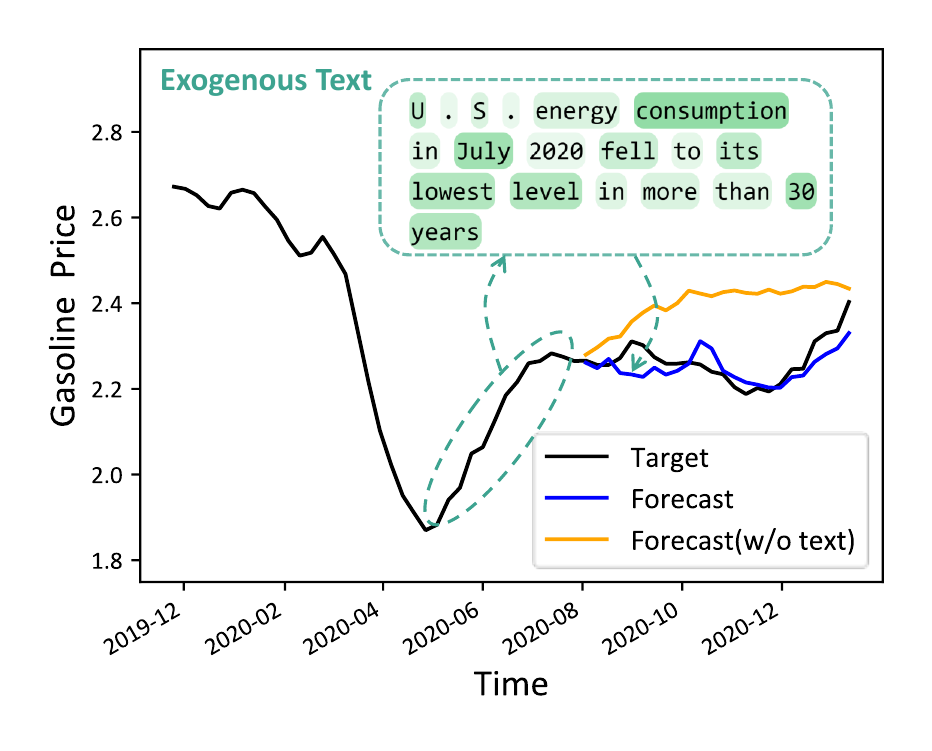}
        \caption{Case study of text abstraction on Energy dataset.}
        \label{sub_case}
    \end{subfigure}
    \label{fig:analysis}
    \caption{Model analysis of CAPTime, including hyperparameter study, model efficiency analysis, and context-aware forecasting visualization.}
\end{figure}

\paragraph{Mixture Distribution Modeling Analysis.} 
To investigate the impact of mixture distribution parameters on performance, specifically the number of experts $M$ and sparse routing parameter $K$, we evaluated six parameter configurations across four datasets. As demonstrated in Fig.~\ref{sub_hyper}, optimal parameter settings vary by dataset scale. For instance, the ETTm1 dataset achieves peak prediction performance with 8 experts and a sparse selection number of 4. Conversely, the smaller Health dataset requires only 4 experts to minimize prediction metrics, indicating that increased modeling complexity does not always improve outcomes.

\paragraph{Efficiency.}
To assess the training efficiency of CAPTime, we conducted experiments on the ETTh1 dataset~\cite{wu2021autoformer}, comparing it with existing LLM-based methods and the high-performance deep forecaster PatchTST~\cite{nie2022time}. The evaluation employed inputs of identical length with a prediction horizon of 336, batch size of 64, using the Adam optimizer, and was conducted on identical hardware. We measured the model's training time (sec/iter), number of trainable parameters (M), and the performance-accuracy trade-off (MSE), as presented in Fig.~\ref{sub_efficiency}. 

CAPTime achieves superior prediction performance with the second-lowest trainable parameter count and the third-fastest training speed. Compared to TimeCMA~\cite{liu2024timecma} and GPT4TS~\cite{zhou2023gpt4ts}, our method exhibits slightly longer training time but significantly improved performance. Comparative evaluation with the top-performing PatchTST reveals that CAPTime outperforms in both efficiency metrics and prediction accuracy. The enhanced efficiency emerges from our model's streamlined architecture which effectively leverages frozen LLM‘s knowledge and benefits from the information compression via text abstraction.

\paragraph{Case Study of Text Abstraction.} 
In the multimodal forecasting dataset~\cite{liu2024timemmd}, the temporal alignment between textual reports and numerical measurements enables clear identification of text's influence on time series variations. Fig.~\ref{sub_case} presents a case study examining text abstraction's impact on forecasting performance in CAPTime. We visualized the attention weights between the final time-series token and exogenous text segments, with darker hues indicating higher attention scores. The results highlight CAPTime's ability to identify correlations, such as declining energy demand leading to reduced gasoline prices. In contrast, the text-agnostic CAPTime (w/o text) fails to detect this correlation, resulting in erroneous prediction trajectories.

\vspace{-0.3cm}

\section{Conclusion}
This paper proposes CAPTime, a multimodal time series forecasting framework that addresses the limitations of existing LLM-based methods in multimodal forecasting. By aligning temporal patterns with text abstractions and incorporating context-aware probabilistic decoding with frozen LLMs, CAPTime demonstrates superior accuracy and generalization capability, particularly in multimodal forecasting scenarios. Moreover, our few-shot and zero-shot results demonstrate that the mixture distribution modeling enhances LLM generalization, while maintaining robustness under data scarcity. CAPTime advances multimodal time series forecasting through text abstraction and probabilistic decoding, enabling applications in healthcare, finance, and traffic modeling. Future work will extend CAPTime to a wider range of multimodal time series tasks.

\bibliographystyle{unsrtnat}
\bibliography{manu}

\begin{thebibliography}{52}
\providecommand{\natexlab}[1]{#1}
\providecommand{\url}[1]{\texttt{#1}}
\expandafter\ifx\csname urlstyle\endcsname\relax
  \providecommand{\doi}[1]{doi: #1}\else
  \providecommand{\doi}{doi: \begingroup \urlstyle{rm}\Url}\fi

\bibitem[Ekambaram et~al.(2024)Ekambaram, Jati, Dayama, Mukherjee, Nguyen, Gifford, Reddy, and Kalagnanam]{ekambaram2024ttms}
Vijay Ekambaram, Arindam Jati, Pankaj Dayama, Sumanta Mukherjee, Nam Nguyen, Wesley~M Gifford, Chandra Reddy, and Jayant Kalagnanam.
\newblock {Tiny Time Mixers (TTMs)}: Fast pre-trained models for enhanced zero/few-shot forecasting of multivariate time series.
\newblock \emph{Advances in Neural Information Processing Systems}, 37:\penalty0 74147--74181, 2024.

\bibitem[Goswami et~al.(2024)Goswami, Szafer, Choudhry, Cai, Li, and Dubrawski]{moment}
Mononito Goswami, Konrad Szafer, Arjun Choudhry, Yifu Cai, Shuo Li, and Artur Dubrawski.
\newblock {MOMENT}: A family of open time-series foundation models.
\newblock In \emph{International Conference on Machine Learning}, pages 16115--16152. PMLR, 2024.

\bibitem[Box et~al.(2015)Box, Jenkins, Reinsel, and Ljung]{box2015time}
George~EP Box, Gwilym~M Jenkins, Gregory~C Reinsel, and Greta~M Ljung.
\newblock \emph{Time series analysis: Forecasting and control}.
\newblock John Wiley \& Sons, 2015.

\bibitem[Sapankevych and Sankar(2009)]{sapankevych2009time}
Nicholas~I Sapankevych and Ravi Sankar.
\newblock Time series prediction using support vector machines: A survey.
\newblock \emph{IEEE Computational Intelligence Magazine}, 4\penalty0 (2):\penalty0 24--38, 2009.

\bibitem[Hochreiter and Schmidhuber(1997)]{hochreiter1997long}
Sepp Hochreiter and J{\"u}rgen Schmidhuber.
\newblock Long short-term memory.
\newblock \emph{Neural Computation}, 9\penalty0 (8):\penalty0 1735--1780, 1997.

\bibitem[Qin et~al.(2017)Qin, Song, Cheng, Cheng, Jiang, and Cottrell]{qin2017dual}
Yao Qin, Dongjin Song, Haifeng Cheng, Wei Cheng, Guofei Jiang, and Garrison~W Cottrell.
\newblock A dual-stage attention-based recurrent neural network for time series prediction.
\newblock In \emph{Proceedings of the 26th International Joint Conference on Artificial Intelligence}, pages 2627--2633, 2017.

\bibitem[Lai et~al.(2018)Lai, Chang, Yang, and Liu]{lai2018modeling}
Guokun Lai, Wei-Cheng Chang, Yiming Yang, and Hanxiao Liu.
\newblock Modeling long-and short-term temporal patterns with deep neural networks.
\newblock In \emph{The 41st International ACM SIGIR Conference on Research \& Development in Information Retrieval}, pages 95--104, 2018.

\bibitem[Bai et~al.(2018)Bai, Kolter, and Koltun]{bai2018empirical}
Shaojie Bai, J~Zico Kolter, and Vladlen Koltun.
\newblock An empirical evaluation of generic convolutional and recurrent networks for sequence modeling.
\newblock \emph{arXiv preprint arXiv:1803.01271}, 2018.

\bibitem[Wu et~al.(2022)Wu, Hu, Liu, Zhou, Wang, and Long]{wutimesnet}
Haixu Wu, Tengge Hu, Yong Liu, Hang Zhou, Jianmin Wang, and Mingsheng Long.
\newblock {TimesNet}: Temporal 2d-variation modeling for general time series analysis.
\newblock In \emph{The Eleventh International Conference on Learning Representations}, 2022.

\bibitem[Luo and Wang(2024)]{luo2024moderntcn}
Donghao Luo and Xue Wang.
\newblock {ModernTCN}: A modern pure convolution structure for general time series analysis.
\newblock In \emph{The Twelfth International Conference on Learning Representations}, pages 1--43, 2024.

\bibitem[Wen et~al.(2023)Wen, Zhou, Zhang, Chen, Ma, Yan, and Sun]{wen2023transformers}
Qingsong Wen, Tian Zhou, Chaoli Zhang, Weiqi Chen, Ziqing Ma, Junchi Yan, and Liang Sun.
\newblock Transformers in time series: A survey.
\newblock In \emph{Proceedings of the Thirty-Second International Joint Conference on Artificial Intelligence}, pages 6778--6786, 2023.

\bibitem[Zhou et~al.(2021)Zhou, Zhang, Peng, Zhang, Li, Xiong, and Zhang]{zhou2021informer}
Haoyi Zhou, Shanghang Zhang, Jieqi Peng, Shuai Zhang, Jianxin Li, Hui Xiong, and Wancai Zhang.
\newblock Informer: Beyond efficient transformer for long sequence time-series forecasting.
\newblock In \emph{Proceedings of the AAAI Conference on Artificial Intelligence}, volume~35, pages 11106--11115, 2021.

\bibitem[Wu et~al.(2021)Wu, Xu, Wang, and Long]{wu2021autoformer}
Haixu Wu, Jiehui Xu, Jianmin Wang, and Mingsheng Long.
\newblock Autoformer: Decomposition transformers with auto-correlation for long-term series forecasting.
\newblock \emph{Advances in Neural Information Processing Systems}, 34:\penalty0 22419--22430, 2021.

\bibitem[Zhou et~al.(2022)Zhou, Ma, Wen, Wang, Sun, and Jin]{zhou2022fedformer}
Tian Zhou, Ziqing Ma, Qingsong Wen, Xue Wang, Liang Sun, and Rong Jin.
\newblock {FEDformer}: Frequency enhanced decomposed transformer for long-term series forecasting.
\newblock In \emph{International Conference on Machine Learning}, pages 27268--27286. PMLR, 2022.

\bibitem[Liu et~al.(2022)Liu, Wu, Wang, and Long]{liu2022non}
Yong Liu, Haixu Wu, Jianmin Wang, and Mingsheng Long.
\newblock {Non-stationary Transformers}: Exploring the stationarity in time series forecasting.
\newblock \emph{Advances in Neural Information Processing Systems}, 35:\penalty0 9881--9893, 2022.

\bibitem[Challu et~al.(2023)Challu, Olivares, Oreshkin, Ramirez, Canseco, and Dubrawski]{challu2023nhits}
Cristian Challu, Kin~G Olivares, Boris~N Oreshkin, Federico~Garza Ramirez, Max~Mergenthaler Canseco, and Artur Dubrawski.
\newblock {NHITS}: Neural hierarchical interpolation for time series forecasting.
\newblock In \emph{Proceedings of the AAAI Conference on Artificial Intelligence}, volume~37, pages 6989--6997, 2023.

\bibitem[Nie et~al.(2022)Nie, Nguyen, Sinthong, and Kalagnanam]{nie2022time}
Yuqi Nie, Nam~H Nguyen, Phanwadee Sinthong, and Jayant Kalagnanam.
\newblock A time series is worth 64 words: Long-term forecasting with transformers.
\newblock \emph{arXiv preprint arXiv:2211.14730}, 2022.

\bibitem[Liu et~al.(2023)Liu, Hu, Zhang, Wu, Wang, Ma, and Long]{liuitransformer}
Yong Liu, Tengge Hu, Haoran Zhang, Haixu Wu, Shiyu Wang, Lintao Ma, and Mingsheng Long.
\newblock {iTransformer}: Inverted transformers are effective for time series forecasting.
\newblock In \emph{The Twelfth International Conference on Learning Representations}, 2023.

\bibitem[Zeng et~al.(2023)Zeng, Chen, Zhang, and Xu]{zeng2023dlinear}
Ailing Zeng, Muxi Chen, Lei Zhang, and Qiang Xu.
\newblock Are transformers effective for time series forecasting?
\newblock In \emph{Proceedings of the AAAI Conference on Artificial Intelligence}, volume~37, pages 11121--11128, 2023.

\bibitem[Oreshkin et~al.(2020)Oreshkin, Carpov, Chapados, and Bengio]{oreshkinnbeats}
Boris~N Oreshkin, Dmitri Carpov, Nicolas Chapados, and Yoshua Bengio.
\newblock N-beats: Neural basis expansion analysis for interpretable time series forecasting.
\newblock In \emph{International Conference on Learning Representations}, 2020.

\bibitem[Gruver et~al.(2023)Gruver, Finzi, Qiu, and Wilson]{gruver2023llmtime}
Nate Gruver, Marc Finzi, Shikai Qiu, and Andrew~G Wilson.
\newblock Large language models are zero-shot time series forecasters.
\newblock \emph{Advances in Neural Information Processing Systems}, 36:\penalty0 19622--19635, 2023.

\bibitem[Zhou et~al.(2023)Zhou, Niu, Sun, Jin, et~al.]{zhou2023gpt4ts}
Tian Zhou, Peisong Niu, Liang Sun, Rong Jin, et~al.
\newblock {One Fits All}: Power general time series analysis by pretrained lm.
\newblock \emph{Advances in Neural Information Processing Systems}, 36:\penalty0 43322--43355, 2023.

\bibitem[Xue and Salim(2023)]{xue2023promptcast}
Hao Xue and Flora~D Salim.
\newblock {PromptCast}: A new prompt-based learning paradigm for time series forecasting.
\newblock \emph{IEEE Transactions on Knowledge and Data Engineering}, 36\penalty0 (11):\penalty0 6851--6864, 2023.

\bibitem[Cao et~al.(2023)Cao, Jia, Arik, Pfister, Zheng, Ye, and Liu]{caotempo}
Defu Cao, Furong Jia, Sercan~O Arik, Tomas Pfister, Yixiang Zheng, Wen Ye, and Yan Liu.
\newblock {TEMPO}: Prompt-based generative pre-trained transformer for time series forecasting.
\newblock In \emph{The Twelfth International Conference on Learning Representations}, 2023.

\bibitem[Jin et~al.(2023)Jin, Wang, Ma, Chu, Zhang, Shi, Chen, Liang, Li, Pan, et~al.]{jintimellm}
Ming Jin, Shiyu Wang, Lintao Ma, Zhixuan Chu, James~Y Zhang, Xiaoming Shi, Pin-Yu Chen, Yuxuan Liang, Yuan-Fang Li, Shirui Pan, et~al.
\newblock {Time-LLM}: Time series forecasting by reprogramming large language models.
\newblock In \emph{The Twelfth International Conference on Learning Representations}, 2023.

\bibitem[Liu et~al.(2024{\natexlab{a}})Liu, Hu, Li, Diao, Liang, Hooi, and Zimmermann]{liu2024unitime}
Xu~Liu, Junfeng Hu, Yuan Li, Shizhe Diao, Yuxuan Liang, Bryan Hooi, and Roger Zimmermann.
\newblock {UniTime}: A language-empowered unified model for cross-domain time series forecasting.
\newblock In \emph{Proceedings of the ACM Web Conference 2024}, pages 4095--4106, 2024{\natexlab{a}}.

\bibitem[Sun et~al.(2024)Sun, Li, Li, and Hong]{suntest}
Chenxi Sun, Hongyan Li, Yaliang Li, and Shenda Hong.
\newblock {TEST}: Text prototype aligned embedding to activate llm's ability for time series.
\newblock In \emph{The Twelfth International Conference on Learning Representations}, 2024.

\bibitem[Pan et~al.(2024)Pan, Jiang, Garg, Schneider, Nevmyvaka, and Song]{pan2024s2ip}
Zijie Pan, Yushan Jiang, Sahil Garg, Anderson Schneider, Yuriy Nevmyvaka, and Dongjin Song.
\newblock {$S^{2}$IP-LLM}: Semantic space informed prompt learning with llm for time series forecasting.
\newblock In \emph{International Conference on Machine Learning}, pages 39135--39153. PMLR, 2024.

\bibitem[Liu et~al.(2024{\natexlab{b}})Liu, Qin, Huang, Wang, and Long]{liu2024autotimes}
Yong Liu, Guo Qin, Xiangdong Huang, Jianmin Wang, and Mingsheng Long.
\newblock {AutoTimes}: Autoregressive time series forecasters via large language models.
\newblock \emph{Advances in Neural Information Processing Systems}, 37:\penalty0 122154--122184, 2024{\natexlab{b}}.

\bibitem[Liu et~al.(2024{\natexlab{c}})Liu, Guo, Dai, Li, Bao, Ren, Jiang, and Xia]{liu2024calf}
Peiyuan Liu, Hang Guo, Tao Dai, Naiqi Li, Jigang Bao, Xudong Ren, Yong Jiang, and Shu-Tao Xia.
\newblock {CALF}: Aligning llms for time series forecasting via cross-modal fine-tuning.
\newblock \emph{arXiv preprint arXiv:2403.07300}, 2024{\natexlab{c}}.

\bibitem[Liu et~al.(2024{\natexlab{d}})Liu, Xu, Miao, Yang, Zhang, Long, Li, and Zhao]{liu2024timecma}
Chenxi Liu, Qianxiong Xu, Hao Miao, Sun Yang, Lingzheng Zhang, Cheng Long, Ziyue Li, and Rui Zhao.
\newblock {TimeCMA}: Towards llm-empowered time series forecasting via cross-modality alignment.
\newblock \emph{arXiv preprint arXiv:2406.01638}, 2024{\natexlab{d}}.

\bibitem[Hu et~al.(2025)Hu, Li, Zhang, Yan, and Chen]{hu2025context}
Yuxiao Hu, Qian Li, Dongxiao Zhang, Jinyue Yan, and Yuntian Chen.
\newblock {Context-Alignment}: Activating and enhancing llm capabilities in time series.
\newblock \emph{arXiv preprint arXiv:2501.03747}, 2025.

\bibitem[Dong et~al.(2024)Dong, Fan, and Peng]{dong2024fnspid}
Zihan Dong, Xinyu Fan, and Zhiyuan Peng.
\newblock {FNSPID}: A comprehensive financial news dataset in time series.
\newblock In \emph{Proceedings of the 30th ACM SIGKDD Conference on Knowledge Discovery and Data Mining}, pages 4918--4927, 2024.

\bibitem[Xu and Cohen(2018)]{xu2018stock}
Yumo Xu and Shay~B Cohen.
\newblock Stock movement prediction from tweets and historical prices.
\newblock In \emph{Proceedings of the 56th Annual Meeting of the Association for Computational Linguistics (Volume 1: Long Papers)}, pages 1970--1979, 2018.

\bibitem[Wang et~al.(2024)Wang, Feng, Qiu, Gu, and Zhao]{wang2024news}
Xinlei Wang, Maike Feng, Jing Qiu, Jinjin Gu, and Junhua Zhao.
\newblock {From News to Forecast}: Integrating event analysis in llm-based time series forecasting with reflection.
\newblock \emph{Advances in Neural Information Processing Systems}, 37:\penalty0 58118--58153, 2024.

\bibitem[Liu et~al.(2024{\natexlab{e}})Liu, Xu, Zhao, Kong, Prabhakar~Kamarthi, Sasanur, Sharma, Cui, Wen, Zhang, et~al.]{liu2024timemmd}
Haoxin Liu, Shangqing Xu, Zhiyuan Zhao, Lingkai Kong, Harshavardhan Prabhakar~Kamarthi, Aditya Sasanur, Megha Sharma, Jiaming Cui, Qingsong Wen, Chao Zhang, et~al.
\newblock {Time-MMD}: Multi-domain multimodal dataset for time series analysis.
\newblock \emph{Advances in Neural Information Processing Systems}, 37:\penalty0 77888--77933, 2024{\natexlab{e}}.

\bibitem[Lee et~al.(2025)Lee, Yu, Shin, Cheng, and Chen]{lee2025timecap}
Geon Lee, Wenchao Yu, Kijung Shin, Wei Cheng, and Haifeng Chen.
\newblock {TimeCAP}: Learning to contextualize, augment, and predict time series events with large language model agents.
\newblock \emph{arXiv preprint arXiv:2502.11418}, 2025.

\bibitem[Jiang et~al.(2025)Jiang, Yu, Lee, Song, Shin, Cheng, Liu, and Chen]{jiang2025timexl}
Yushan Jiang, Wenchao Yu, Geon Lee, Dongjin Song, Kijung Shin, Wei Cheng, Yanchi Liu, and Haifeng Chen.
\newblock Explainable multi-modal time series prediction with llm-in-the-loop.
\newblock \emph{arXiv preprint arXiv:2503.01013}, 2025.

\bibitem[Makridakis et~al.(2020)Makridakis, Spiliotis, and Assimakopoulos]{makridakis2020m4}
Spyros Makridakis, Evangelos Spiliotis, and Vassilios Assimakopoulos.
\newblock The m4 competition: 100,000 time series and 61 forecasting methods.
\newblock \emph{International Journal of Forecasting}, 36\penalty0 (1):\penalty0 54--74, 2020.

\bibitem[Spiliotis et~al.(2020)Spiliotis, Kouloumos, Assimakopoulos, and Makridakis]{spiliotis2020forecasting}
Evangelos Spiliotis, Andreas Kouloumos, Vassilios Assimakopoulos, and Spyros Makridakis.
\newblock Are forecasting competitions data representative of the reality?
\newblock \emph{International Journal of Forecasting}, 36\penalty0 (1):\penalty0 37--53, 2020.

\bibitem[Lv et~al.(2015)Lv, Duan, Kang, Li, and Wang]{lvTrafficFlowPrediction2014}
Yisheng Lv, Yanjie Duan, Wenwen Kang, Zhengxi Li, and Fei-Yue Wang.
\newblock Traffic flow prediction with big data: A deep learning approach.
\newblock \emph{IEEE Transactions on Intelligent Transportation Systems}, 16\penalty0 (2):\penalty0 865--873, 2015.

\bibitem[Jia et~al.(2024)Jia, Wang, Zheng, Cao, and Liu]{jia2024gpt4mts}
Furong Jia, Kevin Wang, Yixiang Zheng, Defu Cao, and Yan Liu.
\newblock Gpt4mts: Prompt-based large language model for multimodal time-series forecasting.
\newblock In \emph{Proceedings of the AAAI Conference on Artificial Intelligence}, volume~38, pages 23343--23351, 2024.

\bibitem[Rodrigues et~al.(2019)Rodrigues, Markou, and Pereira]{rodrigues2019combining}
Filipe Rodrigues, Ioulia Markou, and Francisco~C Pereira.
\newblock Combining time-series and textual data for taxi demand prediction in event areas: A deep learning approach.
\newblock \emph{Information Fusion}, 49:\penalty0 120--129, 2019.

\bibitem[Bommasani et~al.(2021)Bommasani, Hudson, Adeli, Altman, Arora, von Arx, Bernstein, Bohg, Bosselut, Brunskill, et~al.]{bommasani2021foundation}
Rishi Bommasani, Drew~A Hudson, Ehsan Adeli, Russ Altman, Simran Arora, Sydney von Arx, Michael~S Bernstein, Jeannette Bohg, Antoine Bosselut, Emma Brunskill, et~al.
\newblock On the opportunities and risks of foundation models.
\newblock \emph{arXiv preprint arXiv:2108.07258}, 2021.

\bibitem[Das et~al.(2024)Das, Kong, Sen, and Zhou]{das2024timesfm}
Abhimanyu Das, Weihao Kong, Rajat Sen, and Yichen Zhou.
\newblock A decoder-only foundation model for time-series forecasting.
\newblock In \emph{International Conference on Machine Learning}, pages 10148--10167. PMLR, 2024.

\bibitem[Touvron et~al.(2023)Touvron, Lavril, Izacard, Martinet, Lachaux, Lacroix, Rozi{\`e}re, Goyal, Hambro, Azhar, et~al.]{touvron2023llama}
Hugo Touvron, Thibaut Lavril, Gautier Izacard, Xavier Martinet, Marie-Anne Lachaux, Timoth{\'e}e Lacroix, Baptiste Rozi{\`e}re, Naman Goyal, Eric Hambro, Faisal Azhar, et~al.
\newblock Llama: Open and efficient foundation language models.
\newblock \emph{arXiv preprint arXiv:2302.13971}, 2023.

\bibitem[Yang et~al.(2024)Yang, Yang, Zhang, Hui, Zheng, Yu, Li, Liu, Huang, Wei, et~al.]{yang2024qwen2}
An~Yang, Baosong Yang, Beichen Zhang, Binyuan Hui, Bo~Zheng, Bowen Yu, Chengyuan Li, Dayiheng Liu, Fei Huang, Haoran Wei, et~al.
\newblock Qwen2.5 technical report.
\newblock \emph{arXiv preprint arXiv:2412.15115}, 2024.

\bibitem[Rasul et~al.(2023)Rasul, Ashok, Williams, Ghonia, Bhagwatkar, Khorasani, Bayazi, Adamopoulos, Riachi, Hassen, et~al.]{rasul2023lagllama}
Kashif Rasul, Arjun Ashok, Andrew~Robert Williams, Hena Ghonia, Rishika Bhagwatkar, Arian Khorasani, Mohammad Javad~Darvishi Bayazi, George Adamopoulos, Roland Riachi, Nadhir Hassen, et~al.
\newblock {Lag-Llama}: Towards foundation models for probabilistic time series forecasting.
\newblock \emph{arXiv preprint arXiv:2310.08278}, 2023.

\bibitem[Radford et~al.(2019)Radford, Wu, Child, Luan, Amodei, Sutskever, et~al.]{radford2019gpt2}
Alec Radford, Jeffrey Wu, Rewon Child, David Luan, Dario Amodei, Ilya Sutskever, et~al.
\newblock Language models are unsupervised multitask learners.
\newblock \emph{OpenAI blog}, 1\penalty0 (8):\penalty0 9, 2019.

\bibitem[Radford et~al.(2021)Radford, Kim, Hallacy, Ramesh, Goh, Agarwal, Sastry, Askell, Mishkin, Clark, et~al.]{radford2021clip}
Alec Radford, Jong~Wook Kim, Chris Hallacy, Aditya Ramesh, Gabriel Goh, Sandhini Agarwal, Girish Sastry, Amanda Askell, Pamela Mishkin, Jack Clark, et~al.
\newblock Learning transferable visual models from natural language supervision.
\newblock In \emph{International Conference on Machine Learning}, pages 8748--8763. PMLR, 2021.

\bibitem[Fedus et~al.(2022)Fedus, Zoph, and Shazeer]{fedus2022switch}
William Fedus, Barret Zoph, and Noam Shazeer.
\newblock Switch transformers: Scaling to trillion parameter models with simple and efficient sparsity.
\newblock \emph{Journal of Machine Learning Research}, 23\penalty0 (120):\penalty0 1--39, 2022.

\bibitem[Tan et~al.(2024)Tan, Merrill, Gupta, Althoff, and Hartvigsen]{tan2024llmuseful}
Mingtian Tan, Mike Merrill, Vinayak Gupta, Tim Althoff, and Tom Hartvigsen.
\newblock Are language models actually useful for time series forecasting?
\newblock \emph{Advances in Neural Information Processing Systems}, 37:\penalty0 60162--60191, 2024.

\end{thebibliography}


\end{document}